\documentclass{article}
\usepackage{spconf,amsmath,graphicx}
\usepackage{amssymb}
\usepackage{algorithmic}
\usepackage{algorithm}
\usepackage{booktabs}
\usepackage{multirow}
\usepackage{multicol}
\usepackage{appendix}

\title{Prompt Pool based Class-Incremental Continual Learning for Dialog State Tracking}
\name{Hong Liu$^{1,3,\dag}$, Yucheng Cai$^{1,3,\dag}$, Yuan Zhou $^{1,3,\dag}$\thanks{$^\dag$ Equal contribution.}, Zhijian Ou$^{*,1,3}$\thanks{$^{*}$ Corresponding author (ozj@tsinghua.edu.cn).}, Yi Huang$^{2,3}$, Junlan Feng$^{2,3}$
}
\address{
$^{1}$Speech Processing and Machine Intelligence (SPMI) Lab, Tsinghua University, Beijing, China \\
  $^{2}$China Mobile Research Institute, Beijing, China \\
  $^{3}$Tsinghua University-China Mobile Communications Group Co., Ltd. Joint Institute, Beijing, China
}
\copyrightnotice{979-8-3503-0689-7/23/\$31.00~\copyright2023 IEEE}
\begin{document}
%
\maketitle
\begin{abstract}
Continual learning is crucial for dialog state tracking (DST) in dialog systems, since requirements from users for new functionalities are often encountered.
However, most of existing continual learning methods for DST require task identities during testing, which is a severe limit in real-world applications. In this paper, we aim to address continual learning of DST in the class-incremental scenario (namely the task identity is unknown in testing). Inspired by the recently emerging prompt tuning method that performs well on dialog systems, we propose to use the prompt pool method, where we maintain a pool of key-value paired prompts and select prompts from the pool according to the distance between the dialog history and the prompt keys. The proposed method can automatically identify tasks and select appropriate prompts during testing. 
We conduct experiments on Schema-Guided Dialog dataset (SGD) and another dataset collected from a real-world dialog application.
Experiment results show that the prompt pool method achieves much higher joint goal accuracy than the baseline. After combining with a rehearsal buffer, the model performance can be further improved.
\end{abstract}
%
\section{Introduction}
\label{sec:intro}
\begin{figure}[t]
    \centering
    \includegraphics[width=\linewidth]{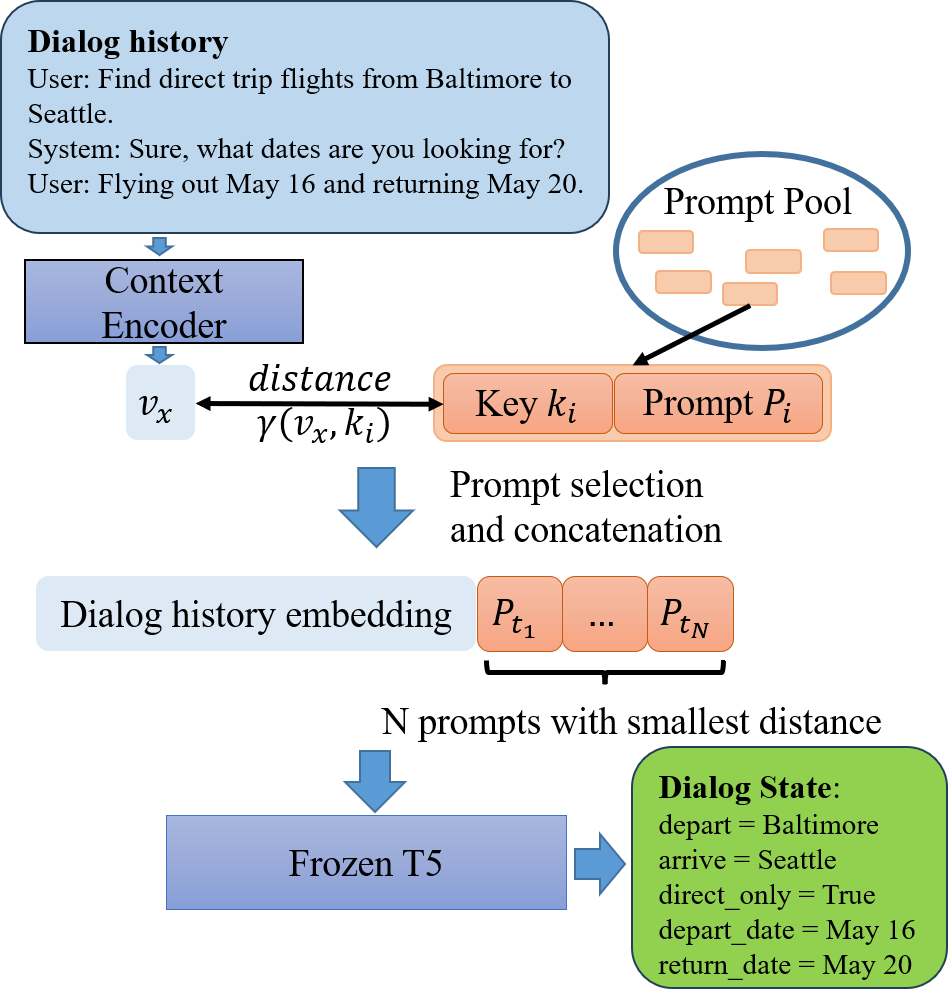}
    \caption{An overview of prompt pool based continual training for DST. Prompt vectors that are close to the context vector will be selected from the pool and concatenated with the dialog history embeddings. The T5 model then takes all the embeddings as input and outputs the dialog state.}
    \label{fig:structure}
    \vspace{-0.5em}
\end{figure}

Task-oriented dialog (TOD) systems are designed to help users accomplish specific goals such as booking flights and finding restaurants.
Dialog state tracking (DST) is an important component in TOD systems, which tracks user goals by inferring structured dialog states expressed in terms of slots and values, as shown in Figure \ref{fig:structure} \cite{williams2016dialog}.
The methods for building DST have been gradually advancing from classification-based \cite{mrkvsic2017neural, dai2018tracking} to sequence-to-sequence generation based \cite{wen2017a, liu2017end, lei2018sequicity, zhang2020task,liu2023building}.
Current DST models are mostly trained in an offline manner, assuming that the domains and required functionalities are fixed through time and that all training data can be accessed beforehand. However, a practical DST model often has to face new tasks. A common requirement is to add new domains for dialogs.
It is costly if the DST model is re-trained when each new task is added. Therefore, continual learning (CL), which refers to expanding a model to new tasks efficiently without forgetting old tasks, i.e., catastrophic forgetting \cite{catastrophic}, is crucial for TOD systems. 

To overcome catastrophic forgetting, three main classes of continual learning algorithms have been developed: \emph{rehearsal}, which  uses an replay buffer to recall previous learned task \cite{de2019episodic,icarl, NIPS2017_f8752278}, \emph{regularization}, which adds regularization term to the loss to avoid forgetting \cite{kirkpatrick2017overcoming, learn_without_forget}, and \emph{architectural}, which trains task-specific component for each task \cite{rusu2022progressive, cl_for_iterative, cl_in_tod, cl_for_dst, l2p}.
However, rehearsal-based methods tend to exhibit decreased performance when the buffer size is reduced, and they are unsuitable for scenarios where the data privacy is a concern. 
Regularization-based methods partially mitigate catastrophic forgetting without the need to store past examples, but they fail to achieve desirable performance in demanding scenarios or intricate datasets \cite{icarl}. 
Architecture-based methods aim to have dedicated components for each task and are more flexible and efficient than the above two classes of methods.
Those task-specific components can be achieved by various approaches such as training a separate adapter \cite{cl_in_tod} or applying network pruning \cite{cl_for_iterative} for each task. However, most architecture-based methods require that the task identity is known during testing. This presents a severe limitation for their application in class-incremental continual learning, for which task identification is necessary at test time.
We leave a short introduction to the three basic scenarios of continual learning \cite{3scenarios}, i.e., task-incremental, domain-incremental and class-incremental, to the section of related work.

In this paper, we are interested in continual learning of DST in the class-incremental learning scenario (namely the task identity is unknown in testing), which is mostly under-explored.
The training data for different domains arrives in a sequence, thus constituting a sequence of tasks.
The DST model needs to be incrementally trained and finally perform well under all tasks.
A recent work in \cite{cl_for_dst} applied continual prompt tuning (CPT) to DST, where it fixes the pretrained language model, trains task-specific prompt vectors for each task, and concatenates those prompts with the context embeddings as the final input embeddings. CPT achieved impressive performance in continual learning of DST. However, CPT cannot work in the class-incremental scenario, because it needs to know the corresponding prompts and slot definitions for the current task before generating dialog states. 

Inspired by the learning to prompt (L2P) method for image classification \cite{l2p}, we propose a prompt pool based continual learning of DST, which can fully support the class-incremental scenario. 
Specifically, we maintain a prompt pool which contains a set of prompt vectors. In addition, each prompt is associated with a key vector for selecting prompts.
For a dialog turn from an arbitrary task, we select prompts from the key-value paired prompt pool according to the distance between the context vector and key vectors. The concatenation of the dialog history embeddings and the selected prompts will be sent into a pretrained model with encoder-decoder structure such as T5 \cite{T5} to predict the dialog state. The parameters of all the prompts and keys will be updated during training, while the context encoder and the pretrained encoder-decoder model are frozen.
The overview of our continual learning framework can be seen in Figure~\ref{fig:structure}.

We conduct experiments of DST on the widely-used Schema-Guided Dialog dataset (SGD) \cite{sgd} and another Chinese dataset collected from a real-world dialog application. We model DST as a sequence-to-sequence generation problem and adjust the sequence format to fit in the class-incremental scenario. The results show that the prompt pool method achieves much higher joint goal accuracy than baseline AdapterCL \cite{cl_in_tod} in class-incremental setting. Moreover, we combine prompt pool with a rehearsal buffer and modify the selection objective for keys, which further improved the model performance. \footnote{ The code is released at https://github.com/thu-spmi/PPT2DST}

\vspace{-1.0em}
\section{Related Work}
\vspace{-1.0em}
\label{sec:related}
\subsection{Continual Learning}
\vspace{-0.5em}
\label{sec:related_cl}
According to training methods, three main classes of continual learning algorithms have been developed: rehearsal-based, regularization-based and architecture-based.
Rehearsal-based methods employ a data buffer to store samples from previous tasks, which are then used for training along with the data from the current task \cite{de2019episodic,icarl, NIPS2017_f8752278}. 
Regularization-based methods restrict the plasticity of the model by constraining the learning rate at important parameters for previous tasks \cite{kirkpatrick2017overcoming, learn_without_forget}.
Architecture-based methods aim to train dedicated components for each task. These task-specific components can be achieved by dynamic expanding network structure \cite{rusu2022progressive}, iteratively applying network pruning and modification\cite{cl_for_iterative}, training a separate adapter \cite{cl_in_tod} or training task-specific prompts \cite{cl_for_dst, l2p} for each task.

According to test scenarios, continual learning can be divided into task-incremental, domain-incremental, and class-incremental \cite{3scenarios}.
Task-incremental learning is the simplest scenario, where the model is aware of the task identity of the current data during testing. Domain-incremental learning is more challenging than task-incremental learning, since the model lacks information about the task identity during testing, but the data labels remain consistent across all tasks, e.g., all tasks are binary classification task. Class-incremental learning is the most complex category among these three, but it is also the closest to real-world scenarios. In class-incremental learning, the model is unaware of the task identity and the labels differ across different tasks.
\vspace{-0.8em}
\subsection{Prompt-based Natural Language Processing}
Recent studies have found that using a textual
prompt can better align pretrained language models to downstream tasks \cite{brown2020gpt3,schick2021exploiting}. Prompt engineering either manually designs prompts \cite{petroni-etal-2019-language} or generates prompts automatically \cite{shin2020autoprompt,gao2021making}. Different from prompt engineering, prompt-tuning adds new tunable prompt tokens to the input, while keep the pretrained model frozen. The added prompts serve as context and affect all following transformer layers, and their embeddings are learned through back-propagation. It is shown that prompt-tuning is parameter-efficient and becomes more competitive with fine-tuning as the model size grows \cite{lester-etal-2021-ptuning}. Prompt-tuning is competitive for continual learning, as only the soft prompts are tuned, instead of the whole pretrained language model. 
Our work proposes to use the prompt pool to leverage the advantage of prompt-tuning, while letting the model to identify tasks and selecting the most appropriate prompts automatically to deal with the class-incremental continual learning problem.

\vspace{-1.0em}
\subsection{Continual Learning in TOD Systems}
\vspace{-0.5em}
Continual learning has been studied in building TOD systems. In \cite{toward_cl}, a regularization-based method, adaptive elastic weight consolidation (AEWC), is utilized to complete continual learning in DST and dialog management. In \cite{cl_in_tod}, separate adapters are trained during continual learning and applied to natural language understanding (NLU), DST and natural language generation (NLG). In \cite{cl_for_nlg}, rehearsal-based and regularization-based methods are combined for NLG in TOD systems. In \cite{cl_for_iterative}, continual learning of NLG is performed by iterative pruning, expanding and masking the network. A recent work in \cite{cl_for_dst} achieve continual learning of DST by applying prompt-tuning \cite{prompt-tuning}, where the pre-trained language model is fixed and the added prompts tokens are tuned to adapt the language model to a sequence of tasks.
However, most of previous studies concentrate on the task-incremental or domain-incremental scenarios, which limites those methods in practical applications. In this work, we aim to address the most challenging class-incremental learning of DST. 

A relevant prior study to our work is AdapterCL \cite{cl_in_tod}, which assumes the task identity is unknown during testing and select the adapter according to the perplexity. However, in \cite{cl_for_dst}, AdapterCL is found to be not parameter-efficient enough, where AdapterCL needs 20 times parameters to catch up with the performance of continual prompt tuning. Though the prompt tuning method has shown its effectiveness in continual learning of DST \cite{cl_for_dst}, it is not suitable for the class-incremental scenario of TOD system.
Our work is inspired by \cite{l2p}, which propose the L2P (also known as prompt pool) method for image classification. The method maintains a prompt pool for continual learning and selects the prompts using key matching. In this work, we further extend the prompt pool method to the class-incremental learning of DST. 

\vspace{-1.0em}

\section{Method}
\vspace{-0.5em}
\label{sec:method}
\subsection{Continual Learning}
In the class-incremental scenario, there are a sequence of tasks $\mathcal{T}=\{\mathcal{T}_1, ..., \mathcal{T}_T\}$. The training data for different tasks arrives in a sequence.
The model needs to be incrementally trained and finally perform well under all tasks.
Denote the corresponding data by $\mathcal{D}=\{\mathcal{D}_1, ..., \mathcal{D}_T\}$, where $\mathcal{D}_t$ denotes the data for the task $\mathcal{T}_t$. $\mathcal{D}_t$ contains multiple data samples $(x^i_t, y^i_t)$ where $x^i_t \in \mathcal{X}_t$ and $y^i_t \in \mathcal{Y}_t$ denote the $i$-th input sample and label. 
We will omit the index $t$ and $i$ in $x^i_t$ later for simplicity, and add a subscript to denote a particular sample such as in Eq. (\ref{eq:x_k}) and (\ref{eq:y_k}) for a particular dialog turn $k$.


\vspace{-1.0em}

\subsection{Dialog State Tracking}
 For dialog state tracking (DST), we need to use the dialog history to predict the dialog state for each turn. The dialog state is a set of slot value pairs: $\{(s_1, v_1), ..., (s_{n_t}, v_{n_t})\}$, where $s$ and $v$ denote slot and value respectively, and $n_t$ is the number of slots used in the task $\mathcal{T}_t$. Usually, all the slots are predefined and DST is to predict the values.
Task-oriented dialogs often consist of dialogs from different domains, such as restaurant service or hotel service. For continual learning of DST, the domain identity is treated as task identity, which is unknown in testing in the class-incremental setting. 

Generally, we formulate DST as text-to-text generation. The DST model accepts an input token sequence and outputs the dialog state, which is also represented by a token sequence. For the input sequence, \cite{cl_for_dst} concatenates slot descriptions and sentinel tokens after dialog history to better predict the value of each slot. However, the task identity is unknown during testing in class-incremental setting, which means that the model does not know which slots are involved in current turn. Therefore, we only take the dialog history as the input sequence, that is
\begin{equation} \label{eq:x_k}
    x_k = u_1 \oplus r_1 \oplus .... \oplus u_{k-1} \oplus r_{k-1} \oplus u_k
\end{equation}
where $u_k, r_k$ denote the user utterance and system response in the $k$-th dialog turn, and $\oplus$ denote the concatenation operation.
For the output sequence, not only the slot values but also the task identity need to be predicted. To simplify the output format, we preset the order of slots in output sequence, so that we only need to predict the values in a specific order during testing. We use special tokens $[s_0], ..., [s_{n_t}]$ to separate values. The empty values are set to be 'none' in the sequence. The output sequence can be formulated as:
\vspace{-0.5em}

\begin{equation} \label{eq:y_k}
    y_k = [s_0]~id_t~[s_1]~v_1^k, ..., [s_{n_t}]~v_{n_t}^k
\end{equation}
where $id_t$ is the identity (i.e., task name) of task $\mathcal{T}_t$. For simplicity, we omit the turn index $k$ in subsequent formulas.
\vspace{-1.0em}
\subsection{Prompt Pool}
\subsubsection{Prompt Tuning}
Prompt tuning \cite{prompt-tuning} simply conditions frozen T5-like language
models \cite{cl_for_dst} to perform down-stream NLP tasks by learning
prompt parameters that are concatenated to the input tokens to instruct the model prediction.
However, ordinary prompt tuning is not applicable to class-incremental learning where the task identity is unknown in testing. 
\cite{l2p} proposed to learn a prompt pool memory space which is served as parameterized instructions for the pretrained model to learn tasks sequentially. The prompt pool is defined as:
\vspace{-0.5em}
\begin{equation}
    \mathbf{P} = \{P_1, P_2, ..., P_J\}
    \vspace{-0.5em}
\end{equation}
where $P_j \in \mathbb{R}^{L_p \times D}$ is a single prompt with token length $L_p$ and embedding size $D$,  and $J$ is the number of prompts in the pool. 
For each task, $N$ prompts, $P_{i_1}, P_{i_2}, ..., P_{i_N}$, will be selected from the prompt pool and concatenated after the embeddings of the dialog history. Let $E(x) \in \mathbb{R}^{|x| \times D}$ denote the embeddings of the input sequence $x$ with token length $|x|$. Then the vector sequence $E_p(x)$ fed into the pretrained model like T5, denoted by $f_\theta(\cdot)$, can be formulated as:
\vspace{-0.5em}
\begin{equation}
\label{eq:input}
    E_p(x) = E(x) \oplus P_{i_1} \oplus ... \oplus P_{i_N}
    \vspace{-0.5em}
\end{equation}
\subsubsection{Prompt Selection}
\label{sec:prompt_select}
To select prompts from the prompt pool, \cite{l2p} model each prompt as a key-value pair: $\{(k_1, P_1), ...,(k_J, P_J)\}$. $k_i \in \mathbb{R}^{D_K}$ is the key for the $i$-th prompt and we denote the set of all keys by $\mathbf{K}$. Intuitively, we can select prompts according to the distance between $x$ and $k_i$. Specifically, the input sequence will be sent into a pretrained encoder model $q_\phi$ to obtain the context vector $c_x \in \mathbb{R}^{D_K}$, for example, the output vector of BERT model corresponding to the '[CLS]' token.
Then the distances between $c_x$ and all the keys will be calculated and $N$ keys with the smallest distances will be selected. The dialog history embeddings and the prompts corresponding to the selected $N$ keys will be concatenated as in Eq.~(\ref{eq:input}).
During training, to ensure that each key in the pool can be selected, we follows \cite{l2p} to directly select $k_{Nt:N(t+1)-1}$ for the task $\mathcal{T}_t$, which is motivated by diversifying prompt selection in \cite{l2p}. 
\vspace{-0.5em}
\subsubsection{Optimization of Prompt Pool}
\vspace{-0.5em}
The optimization of the prompt pool can be divided into two parts. The first part is the cross entropy loss between the model output and the label. The second part is the loss between the context vector and the selected prompt keys.
\vspace{-0.5em}
\begin{equation}
\label{eq:loss}
    \mathcal{L} = CE(f_\theta(E_p(x)), y) + \lambda \sum \limits_{k_i \in \mathbf{K}_x} \gamma(c_x, k_i)
    \vspace{-0.5em}
\end{equation}
where $CE$ denotes the cross entropy loss which is averaged over all tokens in the output sequence, $\gamma$ is a function that calculates Euclid distance and feeds it into a sigmoid function, $\lambda$ is the weight of the second part loss, and $\mathbf{K}_x$ is the set of keys selected from the pool for the input sequence $x$.
The whole training algorithm can be seen in Algorithm~\ref{alg}.
\begin{algorithm}[t]
	\caption{Prompt Pool Training (PPT) for DST}
 	\label{alg}
	\begin{algorithmic}
	    \REQUIRE Frozen pretrained model $f_\theta$, frozen encoder $q_\phi$, task number $T$, a sequence of data $\mathcal{D}$, prompt pool $\mathbf{P}$, prompt keys $\mathbf{K}$, prompt number $N$ for each task, batch size $B$, epochs $E$;
            \STATE Randomly initialize $\mathbf{P}, \mathbf{K}$;
            \FOR{$t$ =1 to $T$}
            \STATE Select $N$ keys and prompts $\mathbf{K}_x=\{K_{Nt}, ..., K_{N(t+1)-1}\}, \mathbf{P}_x=\{P_{Nt}, ..., P_{N(t+1)-1}\}$;
            \FOR{$e$ = 1 to $E$}
		\STATE Obtain a mini-batch of data $\{(x_b, y_b)\}_{r=b}^B$;
            \STATE Calculate the context vector $c_{x_b} = q_\phi(x_b)$ for all input samples;
            \STATE Concatenate the embedding of the input sequence with the selected prompts and obtain an embedding batch $\{(E_p(x_b), y_b)\}_{r=b}^B$;
            \STATE Calculate the loss $\mathcal{L}=$
            \STATE $\frac{1}{B}\sum_{b=1}^B [ CE(f_\theta(E_p(x_b)), y_b) + \lambda \sum \limits_{k_i \in \mathbf{K}_x} \gamma(c_{x_b}, k_i) ]$; 
            \ENDFOR
            \STATE Update the selected keys $\mathbf{K}_x$ and prompts $\mathbf{P}_x$ using the loss $\mathcal{L}$;
            \ENDFOR
	\end{algorithmic}
\end{algorithm}

\vspace{-1.0em}

\subsubsection{Rehearsal Buffer}
Utilizing rehearsal buffer can improve model performance effectively when past data are accessible. For the task $\mathcal{T}_t$, the rehearsal buffer is composed of a fixed number of dialogs selected from previous $t-1$ tasks, which we denote as $M_{<t}$. The new dataset for the task $\mathcal{T}_t$ is $\mathcal{D}_t \cup M_{<t}$.
It is worth noting that the second term in Eq.~(\ref{eq:loss}) is not applicable to the rehearsal-based method, because it will shorten the distance between the context vectors from $M_{<t}$ and keys from $\mathcal{T}_t$.
To address this issue, we change the second term in Eq.~(\ref{eq:loss}) to a binary cross entropy loss. The loss function of the task $\mathcal{T}_t$ can be written as
\begin{equation}
\label{eq:modified_loss}
    \mathcal{L} = CE(f_\theta(E_p(x)), y) + \lambda \sum \limits_{k_i \in \mathbf{K}_x} BCE(\gamma(c_x, k_i), \mathbf{I}(x \in \mathcal{D}_t))
\end{equation}
where $\mathbf{I}(x \in \mathcal{D}_t)$ is an indicator function and $BCE$ is the binary cross entropy loss function, where $BCE(x,y)=-y \log x-(1-y) \log (1-x)$. The algorithm with rehearsal buffer is shown in Algorithm~\ref{alg2} in Appendix.


\section{Experiments}
\label{sec:experiment}
\vspace{-1.5em}
\subsection{Dataset}
\vspace{-0.5em}
We conduct our experiments on Schema-Guided Dialog dataset (SGD) \cite{sgd} that has 44 services over 19 domains. Like \cite{cl_for_dst}, we treat each service as a task and only consider dialogs involving a single service. We randomly select 15 tasks and split the dialogs of one service into train/val/test sets with the ratio of 7:1:2. More details about data statistics can be found in Table~\ref{tab:data_details} in Appendix.
To examine the performance of the method in real-world applications, we conduct experiments on the China Mobile Pickup dataset (CM-Pickup), collected from a real-world dialog application. The purpose is to automatically pickup the incoming call when the phone owner is not available, via dialog state tracking and dialog management. CM-Pickup has 39 domains and we retain 16 domains that have more than 100 dialog sessions and discard other domains.
\vspace{-1.25em}
\subsection{Metrics}
\vspace{-0.5em}
Generally, we evaluate the DST performance using Joint Goal Accuracy (JGA), which calculates the proportion of dialog turns that all the slot values are correctly predicted.
Denote $a_{j, i}$ as the JGA on the test set of the $i$-th task right after training on the $j$-th task. 
A normal measure of the performance of continual learning is the average JGA on all tasks after training on the final task, that is:
\vspace{-0.75em}
\begin{equation}
\vspace{-0.5em}
    JGA_{avg} = \frac{1}{T}\sum\limits_{t=1}^T a_{T, t}
\vspace{-0.25em}
\end{equation}
Besides, we use $f_{t, i}$ to represent the forgetting index of the $i$-th task after training on the task $\mathcal{T}_t$.
\vspace{-0.75em}
\begin{equation}
    f_{t,i} = \max\limits_{j\in[i,t]}a_{j,i}-a_{t, i}
    \vspace{-0.5em}
\end{equation}
We also calculate the accuracy of key selection during testing, which is denoted as $Acc_{key}$. Note that one task corresponds to multiple keys, so counting as correct means that all the keys are selected correctly.
\subsection{Baseline}
\vspace{-0.5em}
We abbreviate the prompt pool training method as PPT and compare it with another class-incremental baseline AdapterCL \cite{cl_in_tod}, which trains a residual adapter for each task and select the adapter with lowest perplexity during testing. For the sake of fairness, we adjust the parameter size of AdapterCL to be close to that of prompt pool.
To improve the performance, we equip PPT with a rehearsal buffer (PPT-R), where we randomly select 50 samples from each task as memory.

We train models using multitask prompt tuning (MPT) method and oracle continual prompt tuning (OCPT) method as the \emph{upper-bound}. The first method trains $N$ prompts using all tasks' data simultaneously. The second method trains $N$ prompts for each task sequentially but the task identity is provided during testing (hence called oracle).
Both are trained using only the first part loss in Eq. (\ref{eq:loss}).
\vspace{-0.5em}
\section{Results}
\vspace{-0.75em}
\subsection{Main Results}
\vspace{-0.5em}
\label{sec:main_result}
Table~\ref{tab:jga} shows the average JGA of the model on 15 tasks after continual learning over SGD. The main findings are as follows: 1) PPT achieves higher $JGA_{avg}$ than AdapterCL; 2) The $JGA_{avg}$ of PPT and PPT-R are still lower than OCPT, which provides task identities in testing. This illustrates the challenge of class-incremental learning; 3) Both PPT and PPT-R have a high accuracy of key selection during testing, indicating that the prompt pool method can be well applied to class-incremental learning scenarios in DST; 4) The rehearsal buffer can improve $JGA_{avg}$ and $Acc_{key}$ effectively.
\begin{table}[t]
    \centering
    \begin{tabular}{l|cc}
    \toprule
         Method & $JGA_{avg}$ & $Acc_{key}$\\
         \hline
         OCPT & 0.481 & -\\
         MPT & 0.614 & -\\
         \hline
         AdapterCL &0.306 & -\\
         PPT &0.346 &0.783\\
         PPT-R &\textbf{0.363} &\textbf{0.811}\\
    \bottomrule
    \end{tabular}
    \caption{Average joint goal accuracy on 15 tasks over SGD. The first block contains the two methods that represent the upper bound and the second shows class-incremental results of different methods.  We also show the key selection accuracy for the PPT methods.}
    \label{tab:jga}
    \vspace{-0.5em}
\end{table}

To better demonstrate how the model performance varies during continual learning, we calculate the forgetting index during training. We only show the first 6 tasks in Table~\ref{tab:forget} due to space limitations. Interestingly, the forgetting index often increases after training on similar tasks. For example, the forgetting index of the second task, flights\_1 increases to 0.176 after training on flights\_3. We speculate that this is because the model cannot distinguish between two similar tasks well (The data distributions of them is close to each other), leading to errors in predicting the task identity of previous tasks.
\begin{table}[t] \footnotesize
    \centering
    \begin{tabular}{l|c}
    \toprule
    Task name & Forgetting index\\
    \hline
    services\_4  &  [0.000, 0.000, 0.000, 0.000, 0.000, 0.000]\\
    flights\_1   & [0.000, 0.000, 0.000, 0.000, 0.000, 0.000]\\
    services\_3 & [\textbf{0.115}, 0.000, 0.000, 0.000, 0.000, 0.000]\\
    flights\_3 & [0.115, \textbf{0.176}, 0.000, 0.000, 0.000, 0.000]\\
    trains\_1 & [0.115, 0.176, 0.000, 0.000, 0.000, 0.000]\\
    homes\_2 &  [0.115, 0.176, 0.000, 0.000, 0.000, 0.000]\\
    \bottomrule
    \end{tabular}
    \caption{The forgetting indices of the first 6 tasks over SGD. Each row contains 6 indices, corresponding to the forgetting index of the current task for the first 6 tasks.}
    \label{tab:forget}
    \vspace{-1.25em}
    
\end{table}

\begin{table}[t] \footnotesize
    \centering
    \begin{tabular}{l|cccc}
    \toprule
         \multirow{2}{*}{Task name} &  \multicolumn{2}{c}{PPT} &\multicolumn{2}{c}{PPT-R}\\
         & JGA &$Acc_{key}$ & JGA &$Acc_{key}$\\
         \hline
         services\_4 &  0.510 &0.793 &0.538 &0.981\\
         flights\_1 & 0.537 &0.741 & 0.563 &0.802\\
         services\_3 &0.534 &1.000 &0.555 &0.990\\
        flights\_3 &0.422 &0.922 &0.353 &0.836\\
        trains\_1 &0.368 &0.983 &0.419 &1.000\\
        homes\_2 &0.144 &0.311 &0.230 &0.547\\
        rentalcars\_2 &0.086 &0.459 &0.416 &0.989\\
        restaurants\_1 &0.497 &1.000 &0.473 &1.000\\
        music\_1 &0.324 &1.000 &0.211 &0.951\\
        hotels\_4 &0.014 &0.014 &0.255 &0.355\\
        media\_2 &0.254 &1.000 &0.338 &0.972\\
        hotels\_3 &0.264 &0.808 &0.228 &0.829\\
        rentalcars\_3 &0.263 &0.828 &0.020 &0.263\\
        hotels\_1 &0.352 &0.880 &0.256 &0.720\\
        homes\_1 &0.628 &1.000 &0.589 &0.930\\
        avg &0.346 &0.783 &0.363 &0.811\\
    \bottomrule
    \end{tabular}
    \caption{The JGA and $Acc_{key}$ of PPT and PPT-R on 15 tasks after continual learning over SGD.}
    \label{tab:details}
    \vspace{-1.0em}
\end{table}

Table~\ref{tab:details} show JGA and $Acc_{key}$ of PPT and PPT-R on different tasks after continual learning. It can be found that JGA and $Acc_{key}$ of most tasks improved after adding a rehearsal buffer. Nonetheless, JGA and $Acc_{key}$ of a few tasks such as rentalcars\_3 have significantly decreased with a rehearsal buffer. After analysis, we found that after adding a rehearsal buffer, the model has a probability of over 70\% misjudging rentalcars\_3 as rentalcars\_1. This phenomenon is consistent with the speculation above. In fact, these similar tasks should be merged into one in a more ideal continuous learning scenario.

\begin{figure}[t]
    \centering
    \includegraphics[width=0.9\linewidth]{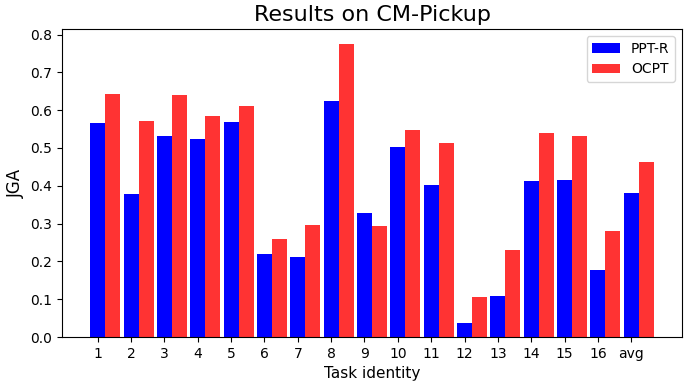}
    \caption{The joint goal accuracy of PPT-R and OCPT (upper-bound) on each task of CM-Pickup.}
    \label{fig:CMresult}
    \vspace{-1.5em}
\end{figure}

For another dataset CM-Pickup, we compare PPT-R with OCPT. The JGA of each task is shown in Figure~\ref{fig:CMresult}. The overall results are similar to Table~\ref{tab:jga}, where PPT-R achieves slightly lower JGA than the upper-bound OOCPT.

\begin{figure}[t]
    \centering
    \includegraphics[width=0.9\linewidth]{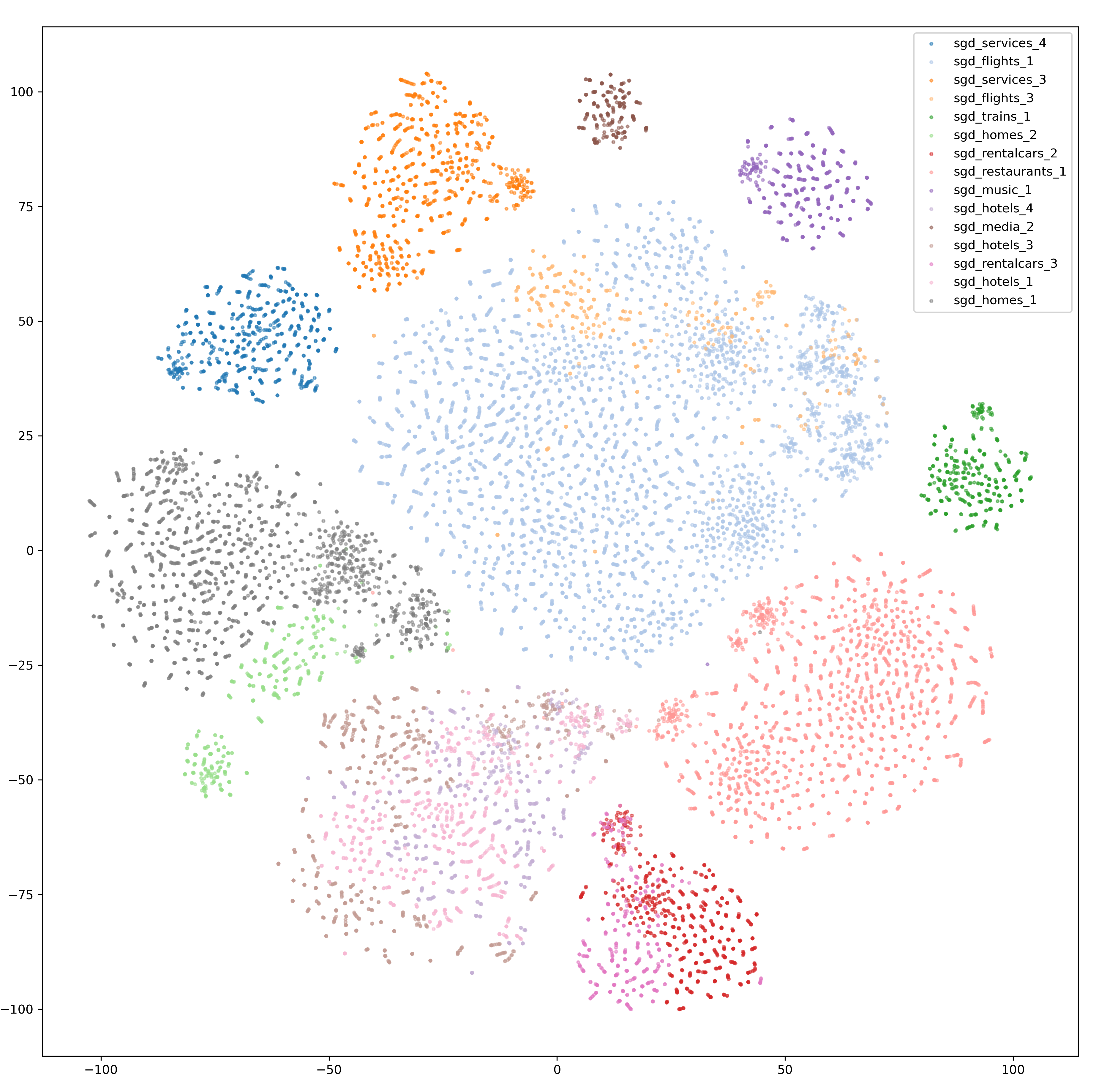}
    \caption{The distribution of context vectors from different tasks in SGD after t-SNE dimension reduction.}
    \label{fig:tsne}
\end{figure}

\begin{table}[t] \small
    \centering
    \begin{tabular}{l|cc}
         \toprule
         Method & $JGA_{avg}$ & $Acc_{key}$\\
         \hline
         PPT-R &\textbf{0.363} &\textbf{0.811}\\
         PPT-R$_{prompt\_only}$ & 0.336 &0.766\\
         PPT-R$_{ordinary}$ & 0.334 & 0.732\\
    \bottomrule
    \end{tabular}
    \caption{Ablation study of the modified key selection loss over SGD.}
    \label{tab:ablation}
    \vspace{-0.5em}
\end{table}


\subsection{Ablation Study}
\vspace{-0.5em}
To understand the similarities between different tasks in SGD, we utilize the t-SNE algorithm \cite{t-SNE} to perform dimension reduction on all the encoded context vectors. The results are shown in Figure~\ref{fig:tsne}. It can be seen that the number of clusters in the figure is less than the number of tasks, 15. For instance, in the middle of the figure, the points of flights\_1 and flights\_3 are closely intertwined.  This indicates that some tasks are similar to each other, which increases the difficulty for the model to distinguish between different tasks.

To reveal the role of the modified loss function, we conduct two ablation experiments on the basis of PPT-R, and the results are shown in Table~\ref{tab:ablation}. The first experiment (PPT-R$_{prompt\_only}$) directly removes the second term in Eq.~(\ref{eq:modified_loss}), that is, the pool of keys, $\mathbf{K}$, is not updated during training. The second (PPT-R$_{ordinary}$) calculates the loss according to Eq.~(\ref{eq:loss}) instead of Eq.~(\ref{eq:modified_loss}) when utilizing rehearsal buffer. Both PPT-R$_{prompt\_only}$ and PPT-R$_{ordinary}$ achieve lower $JGA_{avg}$ and $Acc_{key}$ than PPT-R, and the results of PPT-R$_{ordinary}$ is even lower than those of PPT-R$_{prompt\_only}$. This indicates that the key selection loss in the ordinary loss function in Eq.~(\ref{eq:loss}) is unfavorable for prompt pool training with a rehearsal buffer, and the modified loss in Eq.~(\ref{eq:modified_loss}) can improve the model performance effectively.

To demonstrate that our methods can scale well with the parameters of the backbone, we conduct experiments with different backbones and report the results in Table \ref{tab:scaling}. The results clearly show that our PPT methods scale well with the backbone size. Using a larger model like T5-base and T5-large continually improve the performance of the $JGA_{avg}$ and $Acc_{key}$ metrics. This finding shows that our PPT methods can potentially work well with large language models, which has become prevalent for NLP tasks recently.
\begin{table}[t!]
    \centering
    \begin{tabular}{l|cc}
         \toprule
         Model & $JGA_{avg}$ & $Acc_{key}$\\
         \hline
         T5-small (60M) &0.346 &0.783\\
         T5-base (200M) & 0.420 &0.800\\
         T5-large (750M) & \textbf{0.464} & \textbf{0.822}\\
    \bottomrule
    \end{tabular}
    \caption{Ablation study of the scaling effect of the backbone used for prompt tuning over SGD. PPT is used for all different backbones.
    }
    \label{tab:scaling}
    \vspace{-0.5em}
\end{table}

\vspace{-0.5em}

\section{Conclusion}
\vspace{-0.5em}
\label{sec:conclusion}
We propose to address the class-incremental learning problem of DST using the prompt pool method. We maintain a prompt pool and select the prompts that are close to the input sequence vector during continual learning. The embeddings of the input sequence and the selected prompts will be concatenated together and sent to a pretrained model to predict the dialog state. We conduct experiments on SGD and CM-Pickup and the results show that the prompt pool method outperforms the baseline. We also combine prompt pool with a rehearsal buffer, which further improves the joint goal accuracy. We hope that this work is helpful for building more flexible generative dialog systems for real-world applications.

\clearpage
\bibliographystyle{IEEEbib}
\bibliography{strings,refs}
\clearpage
\appendix
\section{Tables and Algorithms} 
We present the detailed algorithm of prompt pool training with rehearsal buffer, which is shown in Algorithm~\ref{alg2}.
\begin{algorithm}[t]
	\caption{Prompt Pool Training with Rehearsal Buffer (PPT-R) for DST}
 	\label{alg2}
	\begin{algorithmic}
	    \REQUIRE Frozen pretrained model $f_\theta$, frozen encoder $q_\phi$, task number $T$, a sequence of data $\mathcal{D}=\{\mathcal{D}_1, ..., \mathcal{D}_T\}$, prompt pool $\mathbf{P}$, prompt keys $\mathbf{K}$, prompt number $N$ for each task, batch size $B$, epochs $E$, a rehearsal buffer $M$;
            \STATE Randomly initialize $\mathbf{P}, \mathbf{K}$;
            \FOR{$t$ =1 to $T$}
            \STATE Obtain new dataset $\mathcal{D}'_t=\mathcal{D}_t \cup M$;
            \STATE Select $N$ keys and prompts $\mathbf{K}_x=\{K_{Nt}, ..., K_{N(t+1)-1}\}, \mathbf{P}_x=\{P_{Nt}, ..., P_{N(t+1)-1}\}$;
            \FOR{$e$ = 1 to $E$}
		\STATE Obtain a mini-batch of data $\{(x_b, y_b, \mathbf{I}(x_b \in \mathcal{D}_t))\}_{r=b}^B$ from $\mathcal{D}'_t$;
            \STATE Calculate the context vector $c_{x_b} = q_\phi(x_b)$ for all input samples;
            \STATE Concatenate the embedding of the input sequence with the selected prompts and obtain an embedding batch $\{(E_p(x_b), y_b, \mathbf{I}(x_b \in \mathcal{D}_t))\}_{r=b}^B$;
            \STATE Calculate the loss $\mathcal{L}=$
            \STATE $\frac{1}{B}\sum_{b=1}^B [ CE(f_\theta(E_p(x_b)), y_b) + \lambda \sum \limits_{k_i \in \mathbf{K}_x} BCE(\gamma(c_{x_b}, k_i), \mathbf{I}(x_b \in \mathcal{D}_t)) ]$;
            \ENDFOR
            \STATE Update the selected keys $\mathbf{K}_x$ and prompts $\mathbf{P}_x$ using the loss $\mathcal{L}$;
            \STATE Update the rehearsal buffer $M = M \cup S_t$, where $S_t$ denotes 50 dialogs randomly selected from $\mathcal{D}_t$;
            \ENDFOR
	\end{algorithmic}
\end{algorithm}

The statistics of the SGD \cite{sgd} dataset is shown in Table~\ref{tab:data_details}.
\begin{table}[t!]
    \centering
    \begin{tabular}{l|ccc}
    \toprule
         Task name & train & val & test\\
         \hline
         services\_4 & 680 &97 &208\\
         flights\_1 & 4680 &667 &1379\\
         services\_3 &959 &143 &290\\
        flights\_3 &420 &75 &116\\
        trains\_1 &415 &67 &117\\
        homes\_2 &424 &56 &139\\
        rentalcars\_2 &631 &91 &185\\
        restaurants\_1 &2098 &297 &581\\
        music\_1 &468 &73 &142\\
        hotels\_4 &559 &99 &141\\
        media\_2 &215 &29 &71\\
        hotels\_3 &737 &100 &193\\
        rentalcars\_3 &332 &55 &99\\
        hotels\_1 &868 &105 &250\\
        homes\_1 &1829 &282 &540\\
    \bottomrule
    \end{tabular}
    \caption{The number of samples in the 15 selected tasks.}
    \label{tab:data_details}
    \vspace{-1.0em}
\end{table}
\section{Implementation Details}
The pretrained encoder-decoder model $f_\theta$ is a T5-small model \cite{T5}, which has 60M parameters. Besides, we choose Sentence-BERT \cite{sentence-bert} as our sentence encoder model $q_\phi$, which has been found to outperform the T5 encoder in our experiments.
For every task, the epoch number is set to 20, learning rate is set to 0.25 with linear decay to 0, and the weight of the key selection loss $\lambda$ is set to 0.03. As for the prompt pool, the pool size $J=150$ and the number of prompt for each task $N=10$. Each prompt has a token length $L_p=10$. The dimension of keys, $D_K$, is the same as the hidden size of Sentence-BERT, 384, and the dimension of prompts, $D$, is the same as the embedding size of T5-small, 512.

For experiments on Chinese dataset, we choose MT5-small \cite{mt5} as the frozen pretrained model and SBERT-Chinese \cite{uer} as the encoder model.
\end{document}